\documentclass[runningheads]{llncs}

\usepackage{eccv}

\usepackage{eccvabbrv}

\usepackage{graphicx}
\usepackage{booktabs}
\usepackage{multirow}

\usepackage{soul}
\newcommand*\samethanks[1][\value{footnote}]{\footnotemark[#1]}
\usepackage[accsupp]{axessibility}  

\usepackage[pagebackref,breaklinks,colorlinks]{hyperref}

\usepackage{orcidlink}

\begin{document}

\title{General Geometry-aware Weakly Supervised 3D Object Detection} 

\titlerunning{General Geometry-aware Weakly Supervised 3D Object Detection}

\author{Guowen Zhang\inst{1,2}\orcidlink{0000-0001-6692-1185} \and
Junsong Fan\inst{2}\orcidlink{0000-0001-6989-2711} \and
Liyi Chen\inst{1,2}\orcidlink{0000-0001-6600-5064} \and 
Zhaoxiang Zhang \inst{2,3,4}\orcidlink{0000-0003-2648-3875} \and \\
Zhen Lei \inst{2,3,4,\thanks{Corresponding authors. }}\orcidlink{0000-0002-0791-189X} \and
Lei Zhang \inst{1\samethanks}\orcidlink{0000-0002-2078-4215}
}

\authorrunning{G. Zhang et al.}

\institute{The Hong Kong Polytechnic University \\
\email{\{guowen.zhang, liyi.chen\}@connect.polyu.hk, cslzhang@comp.polyu.edu.hk}\\ 
\and Centre for Artificial Intelligence and Robotics, HKISI, CAS\\
\and Institute of Automation, Chinese Academy of Sciences\\
\and School of Artificial Intelligence, University of Chinese Academy of Sciences\\
\email{\{junsong.fan, zhaoxiang.zhang, zlei\}@ia.ac.cn}}

\maketitle

\begin{abstract}
3D object detection is an indispensable component for scene understanding. However, the annotation of large-scale 3D datasets requires significant human effort. To tackle this problem, many methods adopt weakly supervised 3D object detection that estimates 3D boxes by leveraging 2D boxes and scene/class-specific priors. However, these approaches generally depend on sophisticated manual priors, which is hard to generalize to novel categories and scenes. In this paper, we are motivated to propose a general approach, which can be easily adapted to new scenes and/or classes. A unified framework is developed for learning 3D object detectors from RGB images and associated 2D boxes. In specific, we propose three general components: {\textit{prior injection module} to obtain general object geometric priors from LLM model, \textit{2D space projection constraint} to minimize the discrepancy between the boundaries of projected 3D boxes and their corresponding 2D boxes on the image plane, and \textit{3D space geometry constraint} to build a Point-to-Box alignment loss to further refine the pose of estimated 3D boxes.} 
Experiments on KITTI and SUN-RGBD datasets demonstrate that our method yields surprisingly high-quality 3D bounding boxes with only 2D annotation. The source code is available at \href{https://github.com/gwenzhang/GGA}{https://github.com/gwenzhang/GGA}.
\end{abstract}

\section{Introduction}
\label{sec:intro}

%
The primary objective of 3D object detection is to detect tight bounding boxes from point clouds in 3D space, which is essential in many applications, such as autonomous driving~\cite{uniad,world_model_yuqi} and robotics~\cite{LEO}.
Large-scale annotated datasets \cite{Kitti,Nuscenes,Waymo} are critical to training high performance detectors.
However, annotating an adequate number of 3D labels is highly labor-intensive and costly, primarily due to the large workforce of human annotators \cite{huang2019apolloscape, song2015sun}.
%
%
Therefore, seeking efficient proxy annotations for 3D object detection has become attractive~\cite{FGR,AutolabelSDF,WeakM3D,Weankmono3d}.
%
Previous works \cite{tang2019transferable, papadopoulos2017extreme} suggest that labeling 2D bounding boxes is 3 to  16 times faster than 3D bounding boxes.
Additionally, many on-the-shelf large-scale models~\cite{SAM,HQSAM} demonstrate excellent generalization capabilities for 2D objects, which can be a free source for 2D box surrogates. 
Therefore, in this work we are motivated to study how to utilize 2D boxes to train general-purpose 3D object detectors.

\begin{figure}[t]
\centering
\includegraphics[width=0.6\textwidth]{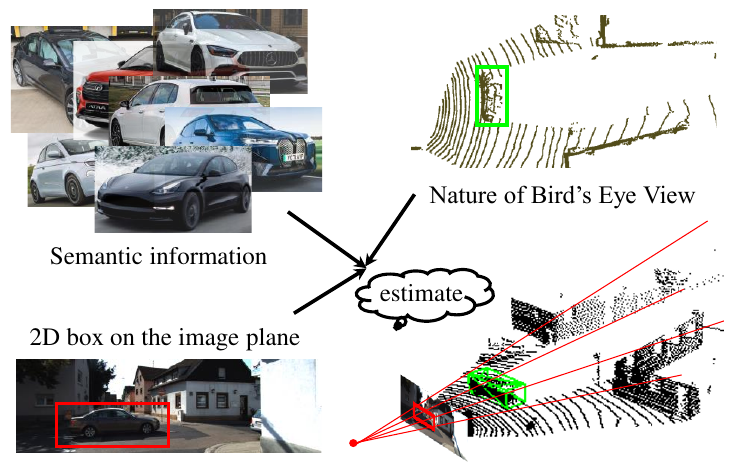}
\caption{\textbf{The insight of our proposed method}. Given RGB-D data, we utilize semantic information, 2D space projection constraints, and 3D space geometry constraints to estimate 3D boxes.}
\label{figure:Insight}
\end{figure}

Some previous efforts attempted to learn 3D detectors from 2D boxes.
For example, geometry-based methods~\cite{WeakM3D,FGR} leverage rule-based information from specific categories in LiDAR point clouds to convert 2D boxes to 3D boxes.
%
Some semi-supervised methods~\cite{WS3D,Mtrans} employed some precisely annotated scenes and instances to regress 3D boxes from 2D boxes.
Besides, some template-based methods~\cite{zakharov2020autolabeling,mccraith2022lifting} adopt SDF~\cite{DeepSDF} or synthetic data~\cite{shapenet} for geometric alignment to convert 2D objects into 3D bounding boxes. 
These methods conduct good practice to use human prior knowledge and intuitive constraints for training. However, their designs are bound to specific object categories and/or model architectures, preventing them from generalizing to different scenes and categories.

To achieve the generalization of weakly supervised 3D detection, in this paper, we propose a General Geometry-Aware (GGA) method, which decomposes the problem into three sub-tasks of \textit{prior injection}, \textit{2D space projection constraints}, and \textit{3D space geometry constraints}, as shown in Fig.~\ref{figure:Insight}. 
The prior injection provides the semantic shape information for the target categories to compensate for the 2D-3D information gap and reduce the data noise. Meanwhile, to ensure universality, it does not rely on any category-specific local geometries.
We consider the ratio information as the prior and propose a semantic ratio loss (SRL) to inject ratio priors.
The ratio is determined by the object's bird-eye view (BEV) width and height.
The category-wise ratio information can be obtained either from humans or Large Language models (LLMs)~\cite{GPT3,llama,gemini}.
While for 2D space projection constraint, we propose a boundary projection loss (BPL) to ensure that the projected wireframe of the estimated 3D boxes aligns closely with the 2D boxes on the image.
However, solely applying constraints on the 2D view space is insufficient for model convergence because there are multiple 3D boxes of different orientations and sizes that can project to the same 2D box. To deal with this problem, we propose a point-to-box alignment loss (PAL), which operates in 3D space and requires the predicted box to tightly wrap the nearby point cluster.

By integrating the three components, we finally arrive at a unified framework to handle the 2D-supervised 3D object detection problem. Our framework is not limited to specific model architectures or scene categories, and it is orthogonal to previous well-tested manual methods, enabling the integration of any category-specific methods. Results on the KITTI~\cite{Kitti} and the SUN-RGBD~\cite{SUN-RGBD} datasets demonstrate that our method is effective and it generalizes well across outdoor and indoor scenarios. In summary, the major contributions of our work are:


\begin{itemize}
    \item We present a novel General Geometry-Aware (GGA) method for weakly supervised 3D object detection from 2D bounding boxes. It can significantly reduce the cost of 3D annotation.
    \item GGA exhibits promising generalization capabilities, allowing it to be easily extended to various novel scenarios and classes.
    \item Extensive experiments on both KITTI and SUN-RGBD benchmarks demonstrate that GGA can generate high-quality 3D boxes from 2D boxes.
\end{itemize}

\section{Related Work}
\subsection{Outdoor 3D Object Detection}
In outdoor scenes such as autonomous driving, point clouds are typically obtained from LiDAR sensors. Those scenes are often large-scale, sparse, and vary in density, which poses stricter requirements for designing efficient detection networks. There are three primary representations used for point cloud learning, point-based, range-view, and voxel-based. Voxel-based detectors \cite{Centerpoint,Voxelnet,SST,Pointpillar,voxel_mamba,scatterformer,msf} employ rasterization to convert point clouds into a volumetric representation, which is used to extract dense features by 3D convolution or Transformers. Point-based methods \cite{Pointnet,Frustumpointnet,Lidarrcnn,Pointrcnn,Pointnetpp} directly extract semantic features from irregular points through a series of downsampling and set abstraction blocks. However, they are limited in the ability to process large-scale regions. Besides, some works convert point clouds into 2D birds-eye-view \cite{Pointpillar} or range view representations \cite{Lasernet,Rangedet,Rangeioudet}, which offer computational benefits compared to voxels. Some hybrid approaches \cite{PVRCNN,PVRCNN++,Point_Voxel_CNN,he2020structure} combine voxel-based and point-based representations to balance the algorithm efficiency and accuracy.

\subsection{Indoor 3D Object Detection}
Different from outdoor scenes, indoor scenarios are characterized by cluttered and restricted environments. Objects are always dense and close to each other, occupying a significant portion of scenes. Therefore, indoor 3D object detection has domain-specific data processing and approaches. Vote-based methods \cite{Votenet,BRNet,H3dnet,Mlcvnet} allocate groups of points to object candidates based on their voted center and extract object features from those point groups. Query-based approaches \cite{3DETR,Groupfree} emerge Transformers for end-to-end training, replacing the voting process with a forward pass during the inference stage. Similar to outdoor scenarios, voxel-based methods \cite{3DSIS,Votenet,Frustum_VoxeNet} convert points into voxel representations and process them using 3D convolution networks.

\subsection{Weakly Supervised 3D Object Detection}
To alleviate heavy 3D annotation costs, many weakly supervised methods and semi-supervised methods have been proposed. WS3D \cite{WS3D} introduces a weakly supervised framework for Lidar-based 3D object detection. It employs center clicks and precisely annotated scenes and instances to regress 3D bounding boxes, resembling a semi-supervised learning setting. Based on the shape prior derived from a pretrained DeepSDF \cite{DeepSDF} on synthetic datasets, SDF \cite{AutolabelSDF} can estimate the 3D geometry of cars detected in 2D images. FGR \cite{FGR} utilizes a non-learning approach to generate pesudo 3D labels by leveraging the specific patterns observed in vehicle point clouds and the frustum. McCraith \textit{et al.} \cite{mccraith2022lifting} combine fixed 3D mesh template and yaw predictors to lift 2D mask to 3D box estimates. 
Weak3DM~\cite{WeakM3D} predicted the 3D boxes from foreground LiDAR points based on hand-craft priors for monocular 3D object detection. Weakmono3D~\cite{Weankmono3d} incorporated projection consistency and multi-view consistency to estimate 3D boxes. Although those methods can lift 2D boxes to 3D box estimates, the sophisticated priors make them hard to extend to novel classes and scenarios.

\section{Methods}

In this section, we first present an overview in Section~\ref{sect::overview}. Then, we introduce the data preparation, backbone, and proposal head of our proposed GGA framework in Section~\ref{sect::GGA}. Finally, we provide a detailed explanation of our proposed BPL, SRL, and PAL from Section~\ref{sect::BPL} to Section~\ref{sect::PAL}.

\subsection{Overview}
\label{sect::overview}
\begin{figure*}[t]
\centering
\includegraphics[width=1.0\textwidth]{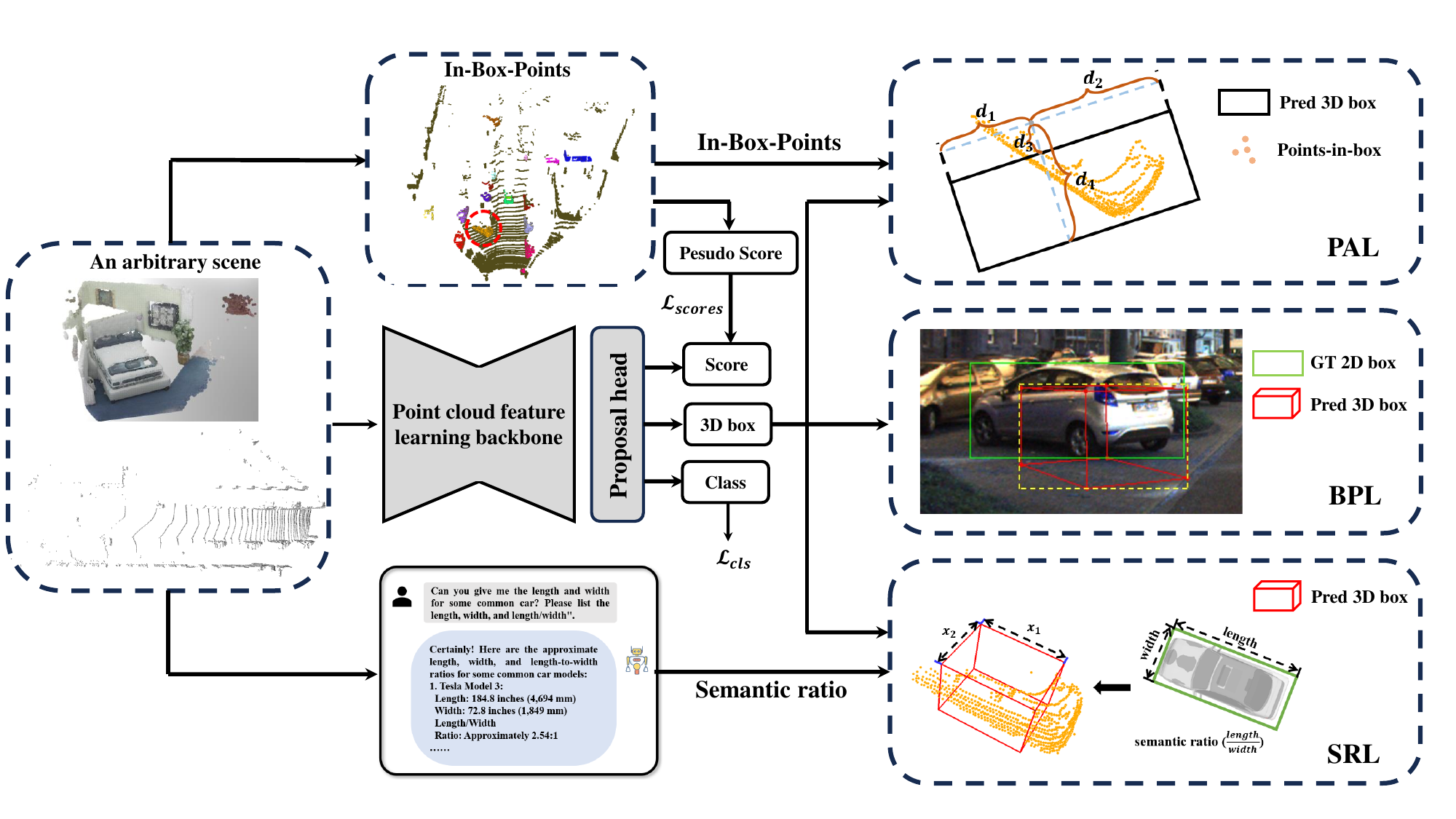}
\caption{\textbf{Illustration of our proposed GGA approach}. Initially, point clouds of a given scene are processed through the point cloud feature learning backbone and proposal head. Meanwhile, we extract the \textit{In-Box-Points} from the 3D frustum corresponding to a 2D bounding box. 
The prior ratio information is obtained from GPT-4 and incorporated into the network through SRL.
The PAL provides the 3D space geometry constraint, while the BPL enforces the 2D space projection constraint.}
\label{figure:Pipeline}
\end{figure*}

An overview of our proposed GGA is shown in Fig.~\ref{figure:Pipeline}.
Given an RGB image and its corresponding point clouds, which can be obtained from LiDAR scans or depth cameras, we aim to classify and localize objects within the scene in 3D space by using 2D bounding boxes $B_{2d}$.
GGA adopts an end-to-end learning-based approach, with the scene point clouds and their corresponding 2D boxes on the image plane as inputs.
For data preprocessing, we follow previous works~\cite{FGR,WeakM3D,Mtrans} to    
select point clouds that can be projected into 2D boxes on the image plane as the initial representation of the corresponding objects, termed \textit{In-Box-Points}. 
Then we estimate the minimum-perimeter 3D pseudo boxes from \textit{In-Box-Points} to provide supervision.
During training, we query the category-specific ratio information from LLMs (i.e., GPT-4~\cite{GPT3}) and inject such priors with the proposed \textit{Semantic Ratio Loss} (SRL). Note that different from previous work~\cite{FGR,AutolabelSDF,mccraith2022lifting,WeakM3D} using manually-collected priors, we fully leverage the LLMs which can provide general cues freely and easily.
Besides, to provide supervision from the 2D perspective, we constrain the geometric relationship between 2D and 3D boxes through a \textit{Boundary Projection Loss} (BPL). To further deal with the ill-posed problem of estimating 3D boxes from 2D boxes, we propose a \textit{Point-to-box Alignment Loss} (PAL) which implicitly supervises 3D boxes and alleviates inherent ambiguity.

%
%
%
%

\subsection{The Proposed GGA Framework}
\label{sect::GGA}
%
%
\textbf{Data Preparation.} 
To obtain the essential components and apply the unified framework of GGA, we first preprocess point clouds of the scene.
Indoor point clouds, compared with outdoor scenarios, are more dense and abundant. To reduce computational complexity, we employ the Farthest Point Sampling~\cite{Pointnet} to reduce raw point clouds to $10^6$ for indoor scenarios.
Besides, following previous works~\cite{FGR,WeakM3D,lsmol}, we remove ground points with the RANSAC algorithm~\cite{RANSAC} to reduce noise in the subsequent \textit{In-Box-Points}.
%
%
%

\textbf{In-Box-Points.} 
Given the Point-Image calibration parameters $P_{proj}$, 3D point clouds can be projected to the image plane.
We select those points within the 2D bounding boxes as the valid points for the corresponding object.
%
Additionally, the selected points often contain outliers that belong to nearby objects and backgrounds.
Following previous methods~\cite{mccraith2022lifting,WeakM3D,FGR}, we use a pre-trained 2D instance segmentation network~\cite{SAM} to filter outliers from valid foreground points in indoor scenarios. The outdoor valid points are denoised by the region growing algorithm~\cite{region_grow}.
%
The resulting point clouds, denoted as \textit{In-Box-Points}: $\mathcal{P}_{ibp}\in\mathbb{R}^{N\times3}$, represent the salient points associated with objects.
Furthermore, we take the minimum-perimeter 3D box that encases all \textit{In-Box-Points} for each object as our initial pseudo box.

\textbf{Backbone.} Recent indoor and outdoor 3D object detection networks consistently utilize domain-specific backbones and heads.
Therefore, our study employs one-stage CenterPoint~\cite{Centerpoint} and FCAF3D~\cite{Fcaf3d} for outdoor and indoor scenarios, respectively.
Both of them are anchor-free methods.
Specifically, CenterPoint uses a 3D encoder to quantize the point clouds into regular bins and then extract features from those voxels.
The backbone utilized in FCAF3D is a sparse adaptation of ResNet~\cite{ResNet}, wherein all 2D convolutions are replaced by sparse 3D convolutions.

\textbf{Proposal head.} 
As shown in Fig.~\ref{figure:Pipeline}, the proposal head mainly outputs the objectness scores, predicted 3D boxes, and object classes.
Due to inherent distinctions, the objectness scores of CenterPoint and FACF3D differ.
CenterPoint generates a K-channel heatmap, with each channel representing one class.
FACF3D, like FCOS~\cite{FCOS}, utilizes centerness to represent the presence of an object at a specific location.
We use initial pseudo boxes to generate pseudo objectness scores, providing the necessary supervision.
For unified subsequent processing, we standardize the predicted 3D box properties to ensure consistency in both indoor and outdoor regressions: a location refinement $({x}_{r}, {y}_{r},{z}_{r}) \in \mathbb{R}^{3}$, a 3D size $(l,w,h) \in \mathbb{R}^{3}$, and a yaw rotation angle $(sin(\alpha), cos(\alpha)) \in \mathbb{R}^{2}$.
For classification, FCAF3D generates classification probability $\hat{p}$, while CenterPoint determines the object class based on class-specific heatmaps.
Finally, by combining the activation position and predicted box properties, we can derive the predicted 3D boxes as:
\begin{equation}
\label{eq::}
B_{3d}^{p}: (x, y, z, l, w, h, sin(\alpha), cos(\alpha)) \in \mathbb{R}^{8},
\end{equation}
where the coordinates $(x, y, z)$ represent the center of the predicted 3D box.
%

\subsection{Boundary Projection Loss}
\label{sect::BPL}
With the Point-Image calibration parameters, a 2D bounding box can be lifted to a frustum, establishing a 3D search space for the object of interest~\cite{Frustumpointnet,FGR}.
This frustum can impose 2D space projection constraints on the predicted 3D boxes, thereby reducing the searching space.
Specifically, in outdoor scenarios, the minimum bounding rectangles of projected 3D bounding boxes on the image plane tend to align closely with the 2D bounding boxes.
%
Inspired by this observation, we propose BPL to locate objects' positions from an image perspective.
In particular, we project the 3D boxes onto the image as follows:
\begin{equation}
\label{eq::project}
(x_{min}^{p}, y_{min}^{p}, x_{max}^{p}, y_{max}^{p}) = \mathbf{Proj}(\mathbf{Corners}(B_{3d}^{p}), P_{proj}),
\end{equation}
where $\mathbf{Corners}(\cdot)$ denotes obtaining the eight corners' coordinates of each predicted 3D box.
Besides, $\mathbf{Proj}(\cdot)$ represents the projection of these corners onto the image by calibration parameters.
$(x_{min}^{p}, y_{min}^{p}, x_{max}^{p}, y_{max}^{p})$ represents the boundaries of the minimum bounding rectangle of those projected corners.
%
%
As shown in Fig.~\ref{figure:Pipeline}, we minimize the discrepancy between the boundaries of the minimum bounding box and the corresponding 2D box as follows:
%
\begin{equation}
\label{eq::BPL}
\begin{split}
\mathcal{L}_\text{BPL} &=L(x_{min}^{p}, x_{min}) + L(x_{max}^{p}, x_{max}) + L(y_{min}^{p}, y_{min}) + L(y_{max}^{p}, y_{max}),
\end{split}
\end{equation}
where $L(\cdot, \cdot)$ is the $L_1$ loss and $(x_{min}, y_{min}, x_{max}, y_{max})$ means the boundaries of $B_{2d}$.
BPL is applied to all the objects of interest and computes their average.
%

%
{Different from outdoor scenarios, the indoor objects are usually closed to the camera, and their images are captured with diverse camera poses, resulting in a significant alignment gap between projected 3D boxes and   2D boxes.} 
%
To address the problem, we project initial pseudo boxes obtained from \textit{In-Box-Points} onto the image plane to obtain the pseudo 2D box.
Then the minimum bounding rectangle that encompasses the pseudo and original 2D box servers as the new 2D constraint.
%
%
%
Overall, we have addressed this issue using a simple yet effective approach, without resorting to any additional information.
\subsection{Semantic Ratio Loss}
\label{sect::SRL}
While the BPL can offer 2D space projection constraints for predicted 3D boxes, it remains challenging to accurately perceive the shape of objects.
Previous methods commonly rely on synthetic data~\cite{AutolabelSDF,mccraith2022lifting} or category-specific local geometries~\cite{FGR,WeakM3D} as shape priors to help learning semantic information.
However, integrating sophisticated hand-craft rules brings challenges for generalization.
%
We observe that fine-grained priors in the network are unnecessary; basic ratio information is sufficient to help the model capture object shapes.
Therefore, we propose a semantic ratio loss to exploit the ratio information for model optimization.
In detail, we take the off-the-shelf GPT-4 to get the required information, noting that large language models contain countless relevant knowledge~\cite{llama,GPT3,gemini}. 
%
Given the predicted 3D box parameters of $(x, y, z, l, w, h, \alpha)$, we aim to apply SRL to supervise the objects' bird-eye view width and height, \ie, $\frac{l}{w}$.
We find that compared to explicitly specifying the output meaning, taking the shorter one as width while another as height achieves better performance. Particularly,
we obtain the predicted ratio with $\frac{min(l, w)}{max(l,w)}$.
SRL is formulated as:
\begin{equation}
\label{eq::SRL}
\mathcal{L}_\text{SRL} = L(\frac{min(l, w)}{max(l,w)}, r),
\end{equation}
where $r$ is the average ratio of various examples generated from GPT-4 and $L(\cdot, \cdot)$ represents the $L_1$ loss. With the prior information from SRL, GGA can converge faster and achieve better performance. 

%

\subsection{Points-to-Box Alignment Loss}
\label{sect::PAL}
A 3D box can be parameterized by multiple properties, such as location, dimension, and rotation.
Methods following the fully supervised paradigm can easily disentangle these properties with annotations.
However, in a weakly supervised manner, there is no general method available to disentangle those properties.
The inherent ambiguity prevents the network from learning accurate 3D boxes.
For example, if the predicted rotation deviates by $\theta$ degrees from the ground truth, other predicted properties (\eg, length and width) can be adjusted to satisfy the requirements.
%
Hence, we propose to use \textit{In-Box-Points} to add implicit supervision for decoupling relevant information in 3D space.
%

%
%
%
Generally speaking, 3D boxes should cover all the foreground points.
We apply the 3D space geometry supervision to predicted 3D boxes based on the BEV location of $\mathcal{P}_{ibp}$.
As shown in Fig.~\ref{figure:Pipeline}, we first compute the distance $(d_{i}^{1}, d_{i}^{2}, d_{i}^{3}, d_{i}^{4})$ from each point in  $\mathcal{P}_{ibp}$ to the four edges of the predicted 3D in the BEV.
Therefore, the 3D space constraint on $\mathcal{P}_{ibp}$ is formulated as:
\begin{equation}
\label{eq::PAL_loc}
\mathcal{L}_{\text{PAL}_{1}} = \sum_{i=1}^{N} (\sum_{j\in\{1, 2\}} \phi(d_{i}^{j} - \frac{l}{2}) + \sum_{k\in\{2, 3\}} \phi(d_{i}^{k} - \frac{w}{2})),
\end{equation}
where $l,w$ is the predicted length and width and $\phi$ is the activation function ReLU.
In addition, RGB-D data capture only a single view of objects, resulting in point clouds that cover only one side of the objects.
Thus, point clouds tend to be concentrated near the boundaries of 3D boxes in BEV.
Inspired by this observation, we propose using this prior to perform implicit supervision.
We further minimize the shortest distance between each point and the four edges of the predicted box in the BEV as follows:
\begin{equation}
\label{eq::PAL_rot}
\mathcal{L}_{\text{PAL}_{2}} = \sum_{i=1}^{N} \text{min}(d_{i}^{1}, d_{i}^{2}, d_{i}^{3}, d_{i}^{4}).
\end{equation}

\subsection{Network Optimization}
We train the network using a combination of our proposed losses, objectness scores loss $\mathcal{L}_\text{score}$, and cross-entropy loss $\mathcal{L}_\text{cls}$ for classification probability:
\begin{equation}
\label{eq::PAL}
\mathcal{L} = \lambda_{1}\cdot\mathcal{L}_\text{BPL} +\lambda_{2}\cdot \mathcal{L}_\text{SRL} + \lambda_{3}\cdot(\mathcal{L}_{\text{PAL}_{1}} + \mathcal{L}_{\text{PAL}_{2}}) + \lambda_{4}\cdot\mathcal{L}_\text{score} + \lambda_{5}\cdot\mathcal{L}_\text{cls},
\end{equation}
where $\lambda_{1-5}$ are the loss weight parameters. $\mathcal{L}_\text{score}$ denotes the heatmap regression loss in CenterPoint and centerness loss in FCAF3D.
Finally, following the common protocol of weakly supervised learning~\cite{WeakM3D,FGR,fpr}, we use the predicted 3D boxes as pseudo labels to train the 3D object detection network under a fully supervised setting.

%
%
%
%
%
%
%

\begin{table*}[t]
\small
 \caption{\textbf{Results on the KITTI val detection benchmark for monocular 3D object detection}. Symbol `-' means that result is not available in the original paper. $\dagger$ means that methods are evaluated with
metric $\mathbf{AP}_{40}$. Others are based on $\mathbf{AP}_{11}$.}
 \label{tab:mono_kitti_val}
\centering
\resizebox{0.7\columnwidth}{!}{
 \begin{tabular}{l|c|c c c}
 \hline
\multirow{2}{*}{Methods} & \multirow{2}{*}{Supversion} & \multicolumn{3}{c}{{$\mathbf{AP_{BEV}/AP_{3D}}${\tiny Car(Iou=0.7)}}}\\
~&~& Easy.&Mod.&Hard.\\
\hline
FQNet~\cite{FQNet}&3D Box&9.50/5.98&8.02/5.50&7.71/4.75\\
Deep3DBox~\cite{Deep3Dbox}&3D Box&9.99/5.85&7.71/4.10&5.30/3.84\\
OFTNet~\cite{OFTNet}&3D Box&11.06/4.07&8.79/3.27&8.91/3.29\\
RoI-10D~\cite{ROI-10D}&3D Box&14.50/10.25&9.91/6.39&8.73/6.18\\
MonoDIS~\cite{MonoDIS}&3D Box&24.26/18.05&18.43/14.98&16.95/13.42\\
MonoPSR~\cite{MonoPSR}&3D Box&20.63/12.75&18.67/11.48&14.45/8.59\\
D4LCN~\cite{D4LCN}&3D Box&26.00/19.38&20.73/16.00&17.46/12.94\\
MonoRun~\cite{monorun}&3D Box&- /20.02&- /14.65&- /12.61\\
PatchNet~\cite{PatchNet}&3D Box&32.30/25.76&21.25/17.72&19.04/15.62\\
PGD~\cite{PGD}&3D Box&30.56/24.35&23.67/18.34&20.84/16.90\\
AutoSDF\cite{AutolabelSDF}&2D Box&15.70/1.23&10.52/0.54&-/-\\
WeakM3D~\cite{WeakM3D}&2D Box&24.89/17.06&16.47/11.63&14.09/{11.17}\\
PGD+GGA & 2D Box & {27.70}/{17.71}&{20.39}/{13.37}&{18.04}/10.89\\
\hline
DCD$^\dagger$~\cite{dcd}&3D Box&32.55/23.81&21.50/15.90&18.25/13.21\\
MonoDTR$^\dagger$~\cite{Monodtr}&3D Box&{33.33}/24.52&{25.35}/18.57&{21.68}/15.51\\
MonoDETR$^\dagger$~\cite{monodetr}&3D Box& - /28.84 & - /20.61 & - /16.38 \\
MonoDETR+GGA$^\dagger$ & 2D Box & {30.07}/{21.18} & {21.49}/14.96 & 18.23/12.25\\
\hline
\end{tabular}}
\end{table*}

\begin{table*}[t]
\small
 \caption{\textbf{Results on the KITTI test detection benchmark for monocular 3D object detection}.}
 \label{tab:mono_kitti_test}
\centering
\resizebox{0.7\columnwidth}{!}{
 \begin{tabular}{l|c|c c c}
 \hline
\multirow{2}{*}{Methods} & \multirow{2}{*}{Supversion} & \multicolumn{3}{c}{{$\mathbf{AP_{BEV}/AP_{3D}}${\tiny Car(Iou=0.7)}}}\\
~&~& Easy.&Mod.&Hard.\\
\hline
FQNet~\cite{FQNet}&3D Box&5.40/2.77&3.23/1.51&2.46/1.01\\
GS3D~\cite{gs3d}&3D Box&8.41/4.47&6.08/2.90&4.94/2.47\\
ROI-10D~\cite{ROI-10D}&3D Box&9.78/4.32&4.91/2.02&3.74/1.46\\
MonoDIS~\cite{MonoDIS}&3D Box&17.23/10.37&{13.19}/{7.94}&{11.12}/{6.40}\\
\hline
WeakM3D~\cite{WeakM3D}&2D Box&11.82/5.03&5.66/2.26&4.08/1.63\\
WeakMono3D~\cite{Weankmono3d}&2D Box&12.31/6.98&8.80/4.85&7.81/4.45\\
PGD+GGA&2D Box&{17.42}/{10.42}&10.21/6.08&8.09/4.65\\
\hline
\end{tabular}}
\end{table*}

\begin{table*}[t]
\small
 \caption{\textbf{Results on KITTI test BEV detection benchmark for Lidar-based 3D object detection methods.}}
 \label{tab:lidar_kitti_test_bev}
\centering
\resizebox{0.8\columnwidth}{!}{
 \begin{tabular}{l|c|c c c|c c c|c c c}
 \hline
\multirow{2}{*}{Methods} & \multirow{2}{*}{Supversion} & \multicolumn{3}{c|}{\multirow{1}{*}{Car (Iou=0.7)}}&\multicolumn{3}{c|}{\multirow{1}{*}{Pedestrain (Iou=0.5)}} &\multicolumn{3}{c}{Cyclist (Iou=0.5)}\\
~&~&Easy.&Mod.&Hard.&Easy.&Mod.&Hard.&Easy.&Mod.&Hard.\\
\hline
MV3D\cite{MV3D}&3D Box&86.02&76.90&68.49&-&-&-&-&-&-\\
VoxelNet\cite{Voxelnet}&3D Box&89.35&79.26&77.39&39.48&33.69&31.50&61.22&48.36&44.37\\
Second~\cite{Second}&3D Box&88.07&79.37&77.95&55.10&46.27&44.76&73.67&56.04&48.78\\
F-PointNet~\cite{Frustumpointnet}&3D Box&88.70&84.00&75.33&51.21&44.89&40.23&72.27&56.12&49.01\\
PointPillars~\cite{Pointpillar}&3D Box&88.35&86.10&79.83&52.08&43.53&41.49&75.78&59.07&52.92\\
PointRCNN~\cite{Pointrcnn}&3D Box&89.28&86.04&79.02&49.43&41.78&38.63&81.36&67.23&59.35\\
\hline
PointRCNN+GGA&2D Box&77.25&63.27&54.70&12.07&11.27&10.47&61.06&46.75&41.84\\
\hline
\end{tabular}}
\end{table*}

\begin{table*}[t]
\small
 \caption{\textbf{AP@0.25 scores for 10 object categories from the SUN-RGBD.}}
 \label{tab:sunrgbd}
\centering
\resizebox{0.9\columnwidth}{!}{
 \begin{tabular}{l|c|c c c c c c c c c c|c}
 \hline
Methods&Supervision&bathtub&bed&bkshelf&chair&desk&dresser&nstand&sofa&table&toilet&mAP\\
\hline
DSS\cite{DSS}&3D Box&44.2&78.7&11.9&61.2&20.5&6.4&15.4&53.5&50.3&78.9&42.1\\
VoteNet\cite{Votenet}&3D Box&74.4&83.0&28.8&75.3&22.0&29.8&62.2&64.0&47.3&90.1&57.7\\
H3DNet\cite{H3dnet}&3D Box&73.8&85.6&31.0&76.7&29.6&33.4&65.5&66.5&50.8&88.2&60.1\\
GroupFree\cite{Groupfree}&3D Box&80.0&87.8&32.5&79.4&32.6&36.0&66.7&70.0&53.8&91.1&63.0\\
FCAF3D\cite{Fcaf3d}&3D Box&79.0&88.3&33.3&81.1&34.0&40.1&71.9&69.7&53.0&91.3&64.2\\
\hline
BoxPC~\cite{BoxPC}&2D Box&29.5&60.9&6.7&36.0&20.2&27.3&50.9&46.4&28.4&65.3&37.2\\
FCAF3D+GGA&2D Box&55.4&69.9&22.4&59.1&22.5&31.3&59.3&58.9&34.8&71.4&48.5\\
\hline
\end{tabular}}
\end{table*}

\begin{table*}[t]
\small
 \caption{Comparisons of different weakly supervised methods on KITTI val set. Symbol `$\times$' means that the method fails to handle the category.}
 \label{tab:compared_kitti}
\centering
\resizebox{1\columnwidth}{!}{
 \begin{tabular}{l|c|c| c| c|c c c|c c c}
 \hline
\multirow{2}{*}{Input} & \multirow{2}{*}{Methods} & \multicolumn{3}{c|}{\multirow{1}{*}{${\text{AP}_\text{BEV}/\text{AP}_\text{3D}}${\tiny Car(Iou=0.5)}}}&\multicolumn{3}{c|}{\multirow{1}{*}{${\text{AP}_\text{BEV}/\text{AP}_\text{3D}}${\tiny Cyclist(Iou=0.25)}}} &\multicolumn{3}{c}{${\text{AP}_\text{BEV}/\text{AP}_\text{3D}}${\tiny Pedestrian(Iou=0.25)}}\\
~&~&\multicolumn{1}{c}{Easy}&\multicolumn{1}{c}{Mod.}&\multicolumn{1}{c|}{Hard}&\multicolumn{1}{c}{Easy}&\multicolumn{1}{c}{Mod.}&\multicolumn{1}{c|}{Hard}&\multicolumn{1}{c}{Easy}&\multicolumn{1}{c}{Mod.}&\multicolumn{1}{c}{Hard}\\
\hline
\multirow{5}{*}{Image}~&VS3D\cite{VS3D}&31.59/22.62&20.59/14.43&16.28/10.91&$\times$&$\times$&$\times$&$\times$&$\times$&$\times$\\
~&AutoSDF~\cite{AutolabelSDF}&50.51/38.31&30.97/19.90&23.72/14.83&$\times$&$\times$&$\times$&$\times$&$\times$&$\times$\\
~&WeakM3D~\cite{WeakM3D}&58.20/50.16&38.02/29.94&30.17/23.11&$\times$&$\times$&$\times$&$\times$&$\times$&$\times$\\
~&WeakMono3D~\cite{Weankmono3d}&54.32/49.37&42.83/39.01&\textbf{40.07/36.34}&-&-&-&-&-&-\\
~&GGA&\textbf{61.45/53.49}&\textbf{43.95/39.50}&37.09/34.22&25.03/23.68&15.77/14.87&14.35/13.99&10.96/9.98&9.76/8.75&8.49/7.87\\
\hline
\multirow{4}{*}{LiDAR}&VS3D~\cite{VS3D}&74.50/40.32&66.71/37.36&57.55/31.09&$\times$&$\times$&$\times$&$\times$&$\times$&$\times$\\
~&AutoSDF~\cite{AutolabelSDF}&77.84/62.25&59.75/42.23&-&$\times$&$\times$&$\times$&$\times$&$\times$&$\times$\\
~&Lifting~\cite{mccraith2022lifting}&86.52/83.45&86.22/79.53&75.53/71.01&$\times$&$\times$&$\times$&$\times$&$\times$&$\times$\\
~&GGA&\textbf{95.25/94.83}&\textbf{87.73/85.19}&\textbf{78.84/76.18}&88.10/86.59&65.68/65.34&61.20/60.78&66.86/65.06&65.61/63.68&61.88/60.01\\
\hline
\end{tabular}}
\end{table*}

\begin{figure}[t]
\centering
\includegraphics[width=0.7\textwidth]{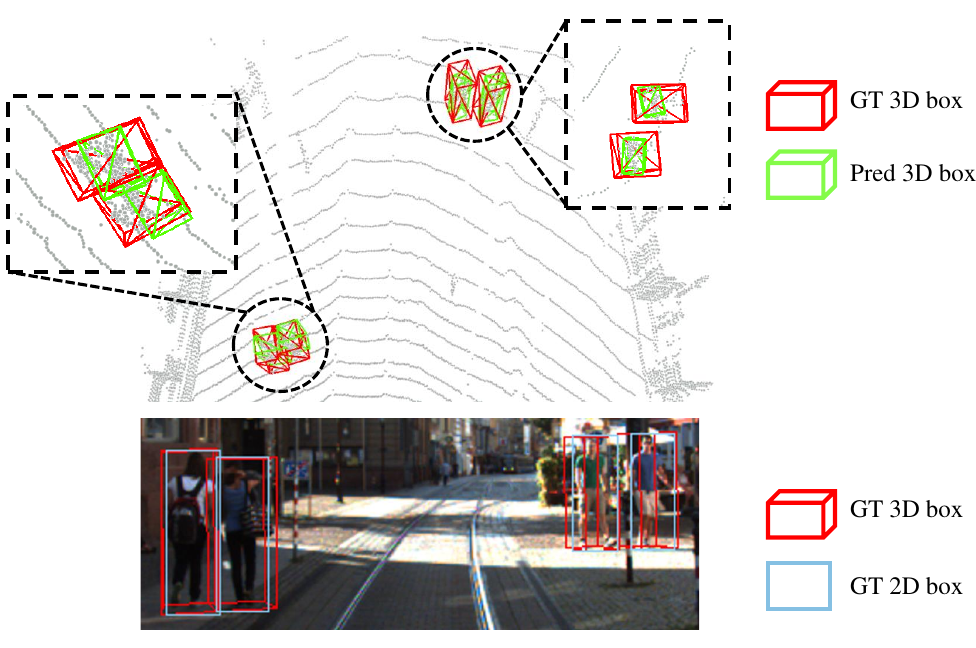}
\caption{\textbf{Generated pseudo 3D boxes of pedestrians by our proposed GGA.}}
\label{figure:Pedestrian_detail}
\end{figure}

\section{Experiments}
\subsection{Experimental Setup}
\textbf{Dataset and evaluation metric.} We conduct the monocular 3D object detection, Lidar-based 3D object detection, and indoor 3D object detection experiments on the SUN-RGBD~\cite{SUN-RGBD} and KITTI~\cite{Kitti} benchmarks.
SUN-RGBD benchmark contains more than 10,000 indoor RGB-D images.
The training and validation splits contain 5,285 and 5,050 point cloud scenes.
Following previous work~\cite{Fcaf3d}, we evaluate on 10 classes.
The KITTI dataset includes 3,712 images for training, 3,769 images for validation, and 7,518 images for testing.
The KITTI benchmark provides evaluations for cars, pedestrians and cyclists.
The evaluation metric is the mean average precision (mAP) with 3D intersection-over-union (IOU) of 0.25 for SUN-RGBD and 0.25/0.5/0.7 for KITTI.

\textbf{Implementation details.} In our experiments, we adopt the CenterPoint~\cite{Centerpoint} and FCAF3D~\cite{Fcaf3d} as the backbone for indoor and outdoor scenarios, respectively.
We implement our method with mmdetection3d~\cite{mmdet3d} and train it with the AdamW optimizer.
The RANSAC threshold is set to 0.04 and 0.2 for indoor and outdoor scenarios, respectively.
The five parameters $\lambda_{1-5}$ in the loss function are set to 2e-3, 2e-3, 1e-4, 1 and 1 for SUN-RGBD dataset.
Following CenterPoint, we omit $\mathcal{L}_\text{cls}$ and set $\lambda_{1-4}$ as 0.3, 0.1, 0.1 and 5 for KITTI benchmark.
We train our network for 120 epochs for outdoor scenarios and 12 epochs for indoor scenarios, respectively.
For the Lidar-based and monocular 3D object detection network, we leverage the generated pseudo 3D bounding boxes to train models (PointRCNN~\cite{Pointrcnn}, PGD~\cite{PGD}, and MonoDETR~\cite{monodetr}) to obtain the final detection results. For indoor scenarios, we utilize FCAF3D~\cite{Fcaf3d}.

\begin{table}[t]
\scriptsize
 \caption{\textbf{Ablation study on the three proposed loss functions on the KITTI val set for Lidar-based 3D object detection.}}
 \label{tab:ablation_loss}
\centering
 \begin{tabular}{ccc|c c c|c c c}
 \hline
\multirow{2}{*}{BPL} & \multirow{2}{*}{SRL} & \multirow{2}{*}{PAL} & \multicolumn{3}{c|}{\multirow{1}{*}{Car (Iou=0.7)}}& \multicolumn{3}{c}{\multirow{1}{*}{Car (Iou=0.5)}}\\
~&~&~&Easy.&Mod.&Hard.&Easy.&Mod.&Hard.\\
\hline
$\checkmark$&$\times$&$\times$&7.92&5.74&4.60&66.33&56.11&48.90\\
$\checkmark$&$\times$&$\checkmark$&26.39&17.73&14.21&78.11&64.93&56.96\\
$\checkmark$&$\checkmark$&$\times$&42.21&28.02&27.45&72.94&61.07&54.35\\
$\checkmark$&$\checkmark$&$\checkmark$&\textbf{56.02}&\textbf{40.35}&\textbf{33.15}&\textbf{84.03}&\textbf{67.81}&\textbf{58.73}\\
\hline
\end{tabular}
\end{table}

\begin{table}[t]
\scriptsize
 \caption{\textbf{Ablation study on the semantic ratio on the KITTI val set for Lidar-based 3D object detection.}}
 \label{tab:ablation_ratio}
\centering
 \begin{tabular}{c|c c c|c c c}
 \hline
\multirow{2}{*}{Semantic Ratio} & \multicolumn{3}{c|}{\multirow{1}{*}{Car (Iou=0.7)}}& \multicolumn{3}{c}{\multirow{1}{*}{Car (Iou=0.5)}}\\
~&\multicolumn{1}{c}{Easy}&\multicolumn{1}{c}{Mod.}&\multicolumn{1}{c|}{Hard}&\multicolumn{1}{c}{Easy}&\multicolumn{1}{c}{Mod.}&\multicolumn{1}{c}{Hard}\\
\hline
1.9&47.35&33.82&28.69&78.54&64.08&\textbf{59.65}\\
2.0&46.25&32.82&27.86&78.71&64.17&57.40\\
2.1&44.44&30.74&29.94&73.53&61.72&56.85\\
2.2&46.90&33.34&28.67&76.05&62.31&56.10\\
2.3&53.00&37.21&31.74&80.68&65.69&57.23\\
2.4&\textbf{56.02}&\textbf{40.35}&33.15&\textbf{84.03}&\textbf{67.81}&58.73\\
2.5&51.20&36.56&31.09&78.02&63.30&56.66\\
2.6&51.32&39.31&\textbf{33.33}&76.64&62.98&57.11\\
2.7&54.74&39.74&32.90&77.37&63.70&57.19\\
2.8&44.05&32.13&27.26&74.45&61.84&55.64\\
2.9&42.56&32.81&26.96&70.48&59.38&53.86\\
\hline
\end{tabular}
\end{table}

\subsection{Experimental Results}

\textbf{Outdoor scenarios.} Table~\ref{tab:mono_kitti_val} shows the evaluation results on the KITTI validation set for monocular 3D object detection, using metrics of $\text{AP}_\text{3D}$ and $\text{AP}_\text{BEV}$. 
Our proposed method significantly outperforms other weakly supervised methods.
MonoDETR with our generated pseudo label achieves 21.18/14.96/12.25 on $\text{AP}_\text{3D}$ for the easy, moderate and hard categories, which are +4.12, +3.33, and + 0.31 better than WeakM3D.
Moreover, although our method does not utilize any 3D bounding box annotations, it outperforms several previous state-of-the-art fully supervised methods. 
For instance, our approach outperforms MonoRun~\cite{monorun} by +1.16 and +0.31 for $\text{AP}_\text{3D}$ in the easy and moderate settings. 
In Table~\ref{tab:mono_kitti_test}, we compare the metrics on the test set for cars with other fully supervised and weakly supervised methods.
Our approach clearly outperforms the state-of-the-art weakly supervised methods WeakMono3D on all metrics. Furthermore, our method achieves comparable results with some fully supervised methods. 

Table~\ref{tab:lidar_kitti_test_bev} shows the performance of our approach on Lidar-based 3D object detection for $\text{AP}_\text{BEV}$ on KITTI test set. Compared with images, point clouds can provide more precise location information and achieve better performance. 
Compared to fully supervised methods, the performance gap between the easy and hard categories for GGA is notably larger.
This is because predicting pseudo labels for objects with few foreground points is difficult.
It is worth noting that the performance of pedestrian detection is comparatively lower. The phenomenon occurs because the annotation process of 3D bounding boxes involves inherent ambiguity. While the ground truth 3D annotation is highly accurate on KITTI, it may not present the only possible solution. Additionally, we give the pedestrian pseudo label generated by our GGA approach in Fig.~\ref{figure:Pedestrian_detail}. It can be observed that our predicted 3D boxes tend to be enclosed by 3D ground truth boxes. Compared with the 3D ground truth, our predicted pseudo 3D boxes may be more tightly aligned with the point clouds. It is challenging to determine which type of annotation is better. Besides, we give the results on the metrics $iou=0.25$ for the pedestrian in Table~\ref{tab:compared_kitti}.

\textbf{Indoor scenarios.} For the indoor scenarios, we evaluate our method on the SUN-RGBD validation set, using metrics of $\text{AP}_{0.25}$. Tab.~\ref{tab:sunrgbd} shows that our proposed framework achieves promising results and outperforms prior fully supervised method DSS. Compared with the semi-supervised BoxPC~\cite{BoxPC}, one can see that even without using any auxiliary 3D boxes, our method significantly outperforms it in all categories.

\textbf{Comparison with other weakly supervised methods.} We compare our methods with other weakly supervised methods on both Lidar-based and monocular 3D object detection. As shown in Tab.~\ref{tab:compared_kitti}, other weakly supervised 3D object detection methods are limited to vehicle categories because they rely on sophisticated manual priors.
Although our method is not specifically designed for the car category, GGA still achieves the best results in almost all metrics.
For monocular 3D object detection, GGA's performance is slightly lower than WeakMono3D in the hard samples. 
This is because WeakMono3D takes advantage of 2D direction labels, which are useful for estimating rotation.
Other weakly supervised 3D object detection methods are limited to vehicle categories because they rely on sophisticated manual priors.

\begin{figure}[t]
\centering
\includegraphics[width=0.4\textwidth]{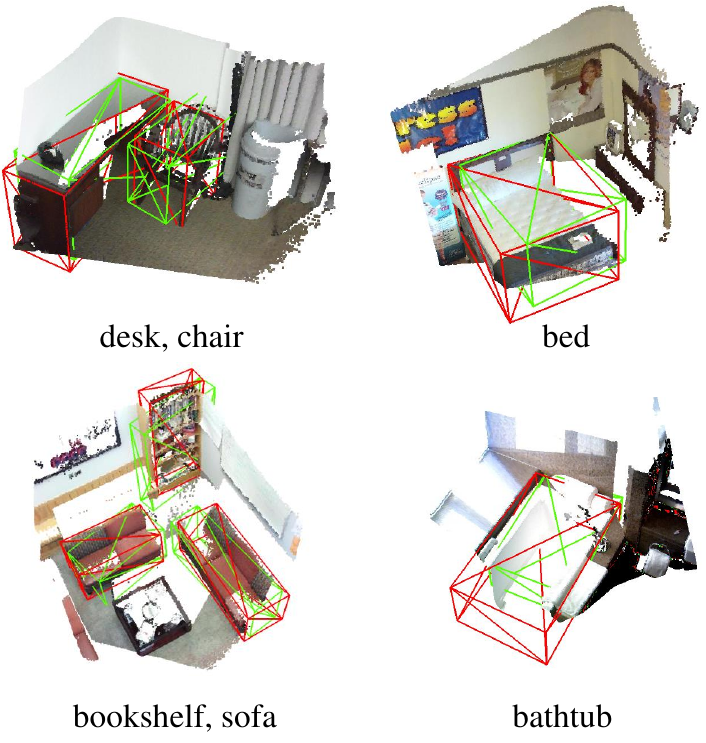}
\caption{\textbf{Visualization of generated 3D boxes for indoor scenarios}. }
\label{figure:indoorbev}
\end{figure}

\begin{figure*}[t]
\centering
\includegraphics[width=0.97\textwidth]{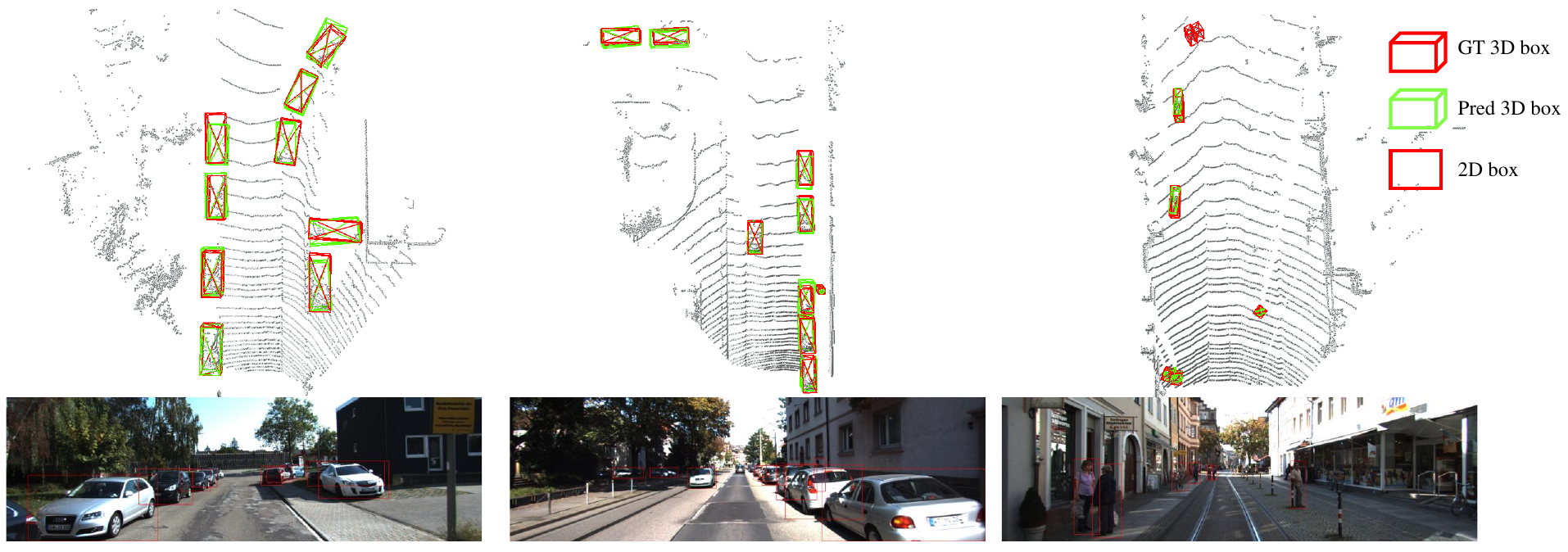}
\caption{\textbf{Visualization of generated 3D boxes in bird's eye view for outdoor scenarios}. }
\label{figure:outdoorbev}
\vspace{-6mm}
\end{figure*}
\subsection{Ablation Study}
To demonstrate the effects of our proposed method, we perform ablation studies for GGA in outdoor scenarios. 
GGA is trained on the KITTI dataset, and we evaluate the quality of the generated pseudo labels against 3D ground truth. In addition, we conduct sensitivity analyses on ratio parameters to demonstrate the robustness of our method.

As shown in Table~\ref{tab:ablation_loss}, we select the backbone with BPL as the initial starting point because we observe that solely relying on SRL or PAL makes it difficult for the model to converge.
The 2D perspective constraint is considered essential for object localization.
The combination of BPL and PAL brings improvements of +18.47 and +9.61 over using only BPL on $\text{AP}_{iou=0.7}$ for the easy and hard samples.
Besides, combining BPL with SRL can bridge the 2D and 3D geometry gap and achieve +34.29 and + 22.85 $\text{AP}_{iou=0.7}$ gains over BPL.
This indicates that semantic prior information is crucial in perceiving the shape of objects.
Finally, utilizing all the proposed losses can achieve the best performance.
PAL provides additional refinement to the predicted 3D boxes by further adjusting their location, rotation and dimensions in conjunction with BPL and SRL.
Compared to using only BPL and PAL, incorporating SRL adds supplementary geometric information and enhances perceptual capabilities.
The ablation study demonstrates the complementary nature of the three proposed losses, resulting in improved 3D box predictions.

We further analyze the effect of the ratio information in Equation~\ref{eq::SRL} on the quality of generated pseudo labels.
In Table~\ref{tab:ablation_ratio}, we present evaluation results of generated pseudo 3D boxes on the KITTI validation set.
GGA achieves optimal results when the semantic ratio is 2.4.
Our proposed method demonstrates a comparable average $\text{AP}_{0.7}$ performance ranging from 39.62 to 43.17 across all difficulty levels when the semantic ratio falls within 2.3 to 2.7, implying that our method exhibits certain robustness to the ratio.

\subsection{Qualitative Results}
Fig.~\ref{figure:indoorbev} illustrates the generated pseudo 3D boxes by GGA in indoor scenarios.
We see that the pseudo 3D boxes are very close to the ground truth 3D boxes.
This indicates that our method can effectively learn the various attributes of 3D bounding boxes.
In Fig.~\ref{figure:outdoorbev}, we visualize the generated pseudo 3D boxes in outdoor scenes.
In the left image, the cars exhibit diverse rotations, yet our method can effectively handle such situations.
Furthermore, most objects in the left and middle images are significantly occluded, and our approach still accurately predicts the 3D boxes in such challenging scenarios.
The right image demonstrates that GGA can generate high-quality 3D boxes for both pedestrians and cyclists.
The visualizations of both indoor and outdoor scenarios validate the generalization ability of GGA across various categories and scenes.

\section{Limitation}
In this paper, we aim to build a bridge between 2D bounding boxes and high-quality 3D boxes for diverse objects.
However, those objects that are far away from the camera contain very few point clouds. The limited point clouds lead to a weak robustness against noise, and it is hard to predict the attributes (\eg, rotation, location, dimension) for 3D boxes with only a few scattered points.
In the future, we plan to explore how to transfer knowledge from objects containing dense points to objects containing very few points.


\section{Conclusion}
In this paper, we focused on the generalization issue for weakly supervised 3D object detection.
Specifically, we proposed a novel general geometry-aware approach to replace the sophisticated hand-craft rules to derive 3D boxes from 2D.
To address the challenge of training on diverse scenes and categories, we indicated that the key is to set up a unified framework for prior injection, 2D space projection constraint and 3D space geometry constraint.
First, we introduced BPL, which serves as a 2D space projection constraint derived from the image plane.
Then, we used ratio information as prior information to help the network to learn semantic knowledge.
Last, we proposed the PAL to refine the box pose in bird's eye view.
Extensive experiments on the KITTI and SUN-RGBD benchmarks demonstrated that our method achieved high performance and it could be easily generalized to novel categories and scenes.

\noindent\textbf{Acknowledgement} This work was supported in part by the InnoHK Program.

%
%
\bibliographystyle{splncs04}
\bibliography{egbib}

\begin{thebibliography}{10}
\providecommand{\url}[1]{\texttt{#1}}
\providecommand{\urlprefix}{URL }
\providecommand{\doi}[1]{https://doi.org/#1}

\bibitem{region_grow}
Adams, R., Bischof, L.: Seeded region growing. IEEE Transactions on pattern analysis and machine intelligence  \textbf{16}(6),  641--647 (1994)

\bibitem{GPT3}
Brown, T., Mann, B., Ryder, N., Subbiah, M., Kaplan, J.D., Dhariwal, P., Neelakantan, A., Shyam, P., Sastry, G., Askell, A., et~al.: Language models are few-shot learners. Advances in neural information processing systems  \textbf{33},  1877--1901 (2020)

\bibitem{Nuscenes}
Caesar, H., Bankiti, V., Lang, A.H., Vora, S., Liong, V.E., Xu, Q., Krishnan, A., Pan, Y., Baldan, G., Beijbom, O.: nuscenes: A multimodal dataset for autonomous driving. In: Proceedings of the IEEE/CVF conference on computer vision and pattern recognition. pp. 11621--11631 (2020)

\bibitem{shapenet}
Chang, A.X., Funkhouser, T., Guibas, L., Hanrahan, P., Huang, Q., Li, Z., Savarese, S., Savva, M., Song, S., Su, H., et~al.: Shapenet: An information-rich 3d model repository. arXiv preprint arXiv:1512.03012  (2015)

\bibitem{monorun}
Chen, H., Huang, Y., Tian, W., Gao, Z., Xiong, L.: Monorun: Monocular 3d object detection by reconstruction and uncertainty propagation. In: Proceedings of the IEEE/CVF Conference on Computer Vision and Pattern Recognition. pp. 10379--10388 (2021)

\bibitem{fpr}
Chen, L., Lei, C., Li, R., Li, S., Zhang, Z., Zhang, L.: Fpr: False positive rectification for weakly supervised semantic segmentation. In: Proceedings of the IEEE/CVF International Conference on Computer Vision. pp. 1108--1118 (2023)

\bibitem{MV3D}
Chen, X., Ma, H., Wan, J., Li, B., Xia, T.: Multi-view 3d object detection network for autonomous driving. In: Proceedings of the IEEE conference on Computer Vision and Pattern Recognition. pp. 1907--1915 (2017)

\bibitem{BRNet}
Cheng, B., Sheng, L., Shi, S., Yang, M., Xu, D.: Back-tracing representative points for voting-based 3d object detection in point clouds. In: Proceedings of the IEEE/CVF conference on computer vision and pattern recognition. pp. 8963--8972 (2021)

\bibitem{mmdet3d}
Contributors, M.: {MMDetection3D: OpenMMLab} next-generation platform for general {3D} object detection. \url{https://github.com/open-mmlab/mmdetection3d} (2020)

\bibitem{D4LCN}
Ding, M., Huo, Y., Yi, H., Wang, Z., Shi, J., Lu, Z., Luo, P.: Learning depth-guided convolutions for monocular 3d object detection. In: Proceedings of the IEEE/CVF Conference on computer vision and pattern recognition workshops. pp. 1000--1001 (2020)

\bibitem{SST}
Fan, L., Pang, Z., Zhang, T., Wang, Y.X., Zhao, H., Wang, F., Wang, N., Zhang, Z.: Embracing single stride 3d object detector with sparse transformer. In: Proceedings of the IEEE/CVF conference on computer vision and pattern recognition. pp. 8458--8468 (2022)

\bibitem{Rangedet}
Fan, L., Xiong, X., Wang, F., Wang, N., Zhang, Z.: Rangedet: In defense of range view for lidar-based 3d object detection. In: Proceedings of the IEEE/CVF international conference on computer vision. pp. 2918--2927 (2021)

\bibitem{RANSAC}
Fischler, M.A., Bolles, R.C.: Random sample consensus: a paradigm for model fitting with applications to image analysis and automated cartography. Communications of the ACM  \textbf{24}(6),  381--395 (1981)

\bibitem{Kitti}
Geiger, A., Lenz, P., Urtasun, R.: Are we ready for autonomous driving? the kitti vision benchmark suite. In: 2012 IEEE conference on computer vision and pattern recognition. pp. 3354--3361. IEEE (2012)

\bibitem{scatterformer}
He, C., Li, R., Zhang, G., Zhang, L.: Scatterformer: Efficient voxel transformer with scattered linear attention. arXiv preprint arXiv:2401.00912  (2024)

\bibitem{msf}
He, C., Li, R., Zhang, Y., Li, S., Zhang, L.: Msf: Motion-guided sequential fusion for efficient 3d object detection from point cloud sequences. In: Proceedings of the IEEE/CVF Conference on Computer Vision and Pattern Recognition. pp. 5196--5205 (2023)

\bibitem{he2020structure}
He, C., Zeng, H., Huang, J., Hua, X.S., Zhang, L.: Structure aware single-stage 3d object detection from point cloud. In: Proceedings of the IEEE/CVF conference on computer vision and pattern recognition. pp. 11873--11882 (2020)

\bibitem{ResNet}
He, K., Zhang, X., Ren, S., Sun, J.: Deep residual learning for image recognition. In: Proceedings of the IEEE conference on computer vision and pattern recognition. pp. 770--778 (2016)

\bibitem{3DSIS}
Hou, J., Dai, A., Nie{\ss}ner, M.: 3d-sis: 3d semantic instance segmentation of rgb-d scans. In: Proceedings of the IEEE/CVF conference on computer vision and pattern recognition. pp. 4421--4430 (2019)

\bibitem{uniad}
Hu, Y., Yang, J., Chen, L., Li, K., Sima, C., Zhu, X., Chai, S., Du, S., Lin, T., Wang, W., et~al.: Planning-oriented autonomous driving. In: Proceedings of the IEEE/CVF Conference on Computer Vision and Pattern Recognition. pp. 17853--17862 (2023)

\bibitem{LEO}
Huang, J., Yong, S., Ma, X., Linghu, X., Li, P., Wang, Y., Li, Q., Zhu, S.C., Jia, B., Huang, S.: An embodied generalist agent in 3d world. arXiv preprint arXiv:2311.12871  (2023)

\bibitem{Monodtr}
Huang, K.C., Wu, T.H., Su, H.T., Hsu, W.H.: Monodtr: Monocular 3d object detection with depth-aware transformer. In: Proceedings of the IEEE/CVF Conference on Computer Vision and Pattern Recognition. pp. 4012--4021 (2022)

\bibitem{huang2019apolloscape}
Huang, X., Wang, P., Cheng, X., Zhou, D., Geng, Q., Yang, R.: The apolloscape open dataset for autonomous driving and its application. IEEE transactions on pattern analysis and machine intelligence  \textbf{42}(10),  2702--2719 (2019)

\bibitem{HQSAM}
Ke, L., Ye, M., Danelljan, M., Liu, Y., Tai, Y.W., Tang, C.K., Yu, F.: Segment anything in high quality. arXiv preprint arXiv:2306.01567  (2023)

\bibitem{SAM}
Kirillov, A., Mintun, E., Ravi, N., Mao, H., Rolland, C., Gustafson, L., Xiao, T., Whitehead, S., Berg, A.C., Lo, W.Y., et~al.: Segment anything. arXiv preprint arXiv:2304.02643  (2023)

\bibitem{MonoPSR}
Ku, J., Pon, A.D., Waslander, S.L.: Monocular 3d object detection leveraging accurate proposals and shape reconstruction. In: Proceedings of the IEEE/CVF conference on computer vision and pattern recognition. pp. 11867--11876 (2019)

\bibitem{Pointpillar}
Lang, A.H., Vora, S., Caesar, H., Zhou, L., Yang, J., Beijbom, O.: Pointpillars: Fast encoders for object detection from point clouds. In: Proceedings of the IEEE/CVF conference on computer vision and pattern recognition. pp. 12697--12705 (2019)

\bibitem{gs3d}
Li, B., Ouyang, W., Sheng, L., Zeng, X., Wang, X.: Gs3d: An efficient 3d object detection framework for autonomous driving. In: Proceedings of the IEEE/CVF Conference on Computer Vision and Pattern Recognition. pp. 1019--1028 (2019)

\bibitem{dcd}
Li, Y., Chen, Y., He, J., Zhang, Z.: Densely constrained depth estimator for monocular 3d object detection. In: European Conference on Computer Vision. pp. 718--734. Springer (2022)

\bibitem{Lidarrcnn}
Li, Z., Wang, F., Wang, N.: Lidar r-cnn: An efficient and universal 3d object detector. In: Proceedings of the IEEE/CVF Conference on Computer Vision and Pattern Recognition. pp. 7546--7555 (2021)

\bibitem{Rangeioudet}
Liang, Z., Zhang, Z., Zhang, M., Zhao, X., Pu, S.: Rangeioudet: Range image based real-time 3d object detector optimized by intersection over union. In: Proceedings of the IEEE/CVF Conference on Computer Vision and Pattern Recognition. pp. 7140--7149 (2021)

\bibitem{Mtrans}
Liu, C., Qian, X., Huang, B., Qi, X., Lam, E., Tan, S.C., Wong, N.: Multimodal transformer for automatic 3d annotation and object detection. In: European Conference on Computer Vision. pp. 657--673. Springer (2022)

\bibitem{FQNet}
Liu, L., Lu, J., Xu, C., Tian, Q., Zhou, J.: Deep fitting degree scoring network for monocular 3d object detection. In: Proceedings of the IEEE/CVF conference on computer vision and pattern recognition. pp. 1057--1066 (2019)

\bibitem{Groupfree}
Liu, Z., Zhang, Z., Cao, Y., Hu, H., Tong, X.: Group-free 3d object detection via transformers. 2021 ieee. In: CVF International Conference on Computer Vision (ICCV). pp. 2929--2938 (2021)

\bibitem{Point_Voxel_CNN}
Liu, Z., Tang, H., Lin, Y., Han, S.: Point-voxel cnn for efficient 3d deep learning. Advances in Neural Information Processing Systems  \textbf{32} (2019)

\bibitem{PatchNet}
Ma, X., Liu, S., Xia, Z., Zhang, H., Zeng, X., Ouyang, W.: Rethinking pseudo-lidar representation. In: Computer Vision--ECCV 2020: 16th European Conference, Glasgow, UK, August 23--28, 2020, Proceedings, Part XIII 16. pp. 311--327. Springer (2020)

\bibitem{ROI-10D}
Manhardt, F., Kehl, W., Gaidon, A.: Roi-10d: Monocular lifting of 2d detection to 6d pose and metric shape. In: Proceedings of the IEEE/CVF Conference on Computer Vision and Pattern Recognition. pp. 2069--2078 (2019)

\bibitem{mccraith2022lifting}
McCraith, R., Insafutdinov, E., Neumann, L., Vedaldi, A.: Lifting 2d object locations to 3d by discounting lidar outliers across objects and views. In: 2022 International Conference on Robotics and Automation (ICRA). pp. 2411--2418. IEEE (2022)

\bibitem{WS3D}
Meng, Q., Wang, W., Zhou, T., Shen, J., Van~Gool, L., Dai, D.: Weakly supervised 3d object detection from lidar point cloud. In: European Conference on computer vision. pp. 515--531. Springer (2020)

\bibitem{Lasernet}
Meyer, G.P., Laddha, A., Kee, E., Vallespi-Gonzalez, C., Wellington, C.K.: Lasernet: An efficient probabilistic 3d object detector for autonomous driving. In: Proceedings of the IEEE/CVF conference on computer vision and pattern recognition. pp. 12677--12686 (2019)

\bibitem{3DETR}
Misra, I., Girdhar, R., Joulin, A.: An end-to-end transformer model for 3d object detection. In: Proceedings of the IEEE/CVF International Conference on Computer Vision. pp. 2906--2917 (2021)

\bibitem{Deep3Dbox}
Mousavian, A., Anguelov, D., Flynn, J., Kosecka, J.: 3d bounding box estimation using deep learning and geometry. In: Proceedings of the IEEE conference on Computer Vision and Pattern Recognition. pp. 7074--7082 (2017)

\bibitem{papadopoulos2017extreme}
Papadopoulos, D.P., Uijlings, J.R., Keller, F., Ferrari, V.: Extreme clicking for efficient object annotation. In: Proceedings of the IEEE international conference on computer vision. pp. 4930--4939 (2017)

\bibitem{DeepSDF}
Park, J.J., Florence, P., Straub, J., Newcombe, R., Lovegrove, S.: Deepsdf: Learning continuous signed distance functions for shape representation. In: Proceedings of the IEEE/CVF conference on computer vision and pattern recognition. pp. 165--174 (2019)

\bibitem{WeakM3D}
Peng, L., Yan, S., Wu, B., Yang, Z., He, X., Cai, D.: Weakm3d: Towards weakly supervised monocular 3d object detection. arXiv preprint arXiv:2203.08332  (2022)

\bibitem{Votenet}
Qi, C.R., Litany, O., He, K., Guibas, L.J.: Deep hough voting for 3d object detection in point clouds. In: proceedings of the IEEE/CVF International Conference on Computer Vision. pp. 9277--9286 (2019)

\bibitem{Frustumpointnet}
Qi, C.R., Liu, W., Wu, C., Su, H., Guibas, L.J.: Frustum pointnets for 3d object detection from rgb-d data. In: Proceedings of the IEEE conference on computer vision and pattern recognition. pp. 918--927 (2018)

\bibitem{Pointnet}
Qi, C.R., Su, H., Mo, K., Guibas, L.J.: Pointnet: Deep learning on point sets for 3d classification and segmentation. In: Proceedings of the IEEE conference on computer vision and pattern recognition. pp. 652--660 (2017)

\bibitem{Pointnetpp}
Qi, C.R., Yi, L., Su, H., Guibas, L.J.: Pointnet++: Deep hierarchical feature learning on point sets in a metric space. Advances in neural information processing systems  \textbf{30} (2017)

\bibitem{VS3D}
Qin, Z., Wang, J., Lu, Y.: Weakly supervised 3d object detection from point clouds. In: Proceedings of the 28th ACM International Conference on Multimedia. pp. 4144--4152 (2020)

\bibitem{OFTNet}
Roddick, T., Kendall, A., Cipolla, R.: Orthographic feature transform for monocular 3d object detection. arXiv preprint arXiv:1811.08188  (2018)

\bibitem{Fcaf3d}
Rukhovich, D., Vorontsova, A., Konushin, A.: Fcaf3d: Fully convolutional anchor-free 3d object detection. In: European Conference on Computer Vision. pp. 477--493. Springer (2022)

\bibitem{Frustum_VoxeNet}
Shen, X., Stamos, I.: Frustum voxnet for 3d object detection from rgb-d or depth images. In: Proceedings of the IEEE/CVF Winter Conference on Applications of Computer Vision. pp. 1698--1706 (2020)

\bibitem{PVRCNN}
Shi, S., Guo, C., Jiang, L., Wang, Z., Shi, J., Wang, X., Li, H.: Pv-rcnn: Point-voxel feature set abstraction for 3d object detection. In: Proceedings of the IEEE/CVF conference on computer vision and pattern recognition. pp. 10529--10538 (2020)

\bibitem{PVRCNN++}
Shi, S., Jiang, L., Deng, J., Wang, Z., Guo, C., Shi, J., Wang, X., Li, H.: Pv-rcnn++: Point-voxel feature set abstraction with local vector representation for 3d object detection. International Journal of Computer Vision  \textbf{131}(2),  531--551 (2023)

\bibitem{Pointrcnn}
Shi, S., Wang, X., Li, H.: Pointrcnn: 3d object proposal generation and detection from point cloud. In: Proceedings of the IEEE/CVF conference on computer vision and pattern recognition. pp. 770--779 (2019)

\bibitem{MonoDIS}
Simonelli, A., Bulo, S.R., Porzi, L., L{\'o}pez-Antequera, M., Kontschieder, P.: Disentangling monocular 3d object detection. In: Proceedings of the IEEE/CVF International Conference on Computer Vision. pp. 1991--1999 (2019)

\bibitem{song2015sun}
Song, S., Lichtenberg, S.P., Xiao, J.: Sun rgb-d: A rgb-d scene understanding benchmark suite. In: Proceedings of the IEEE conference on computer vision and pattern recognition. pp. 567--576 (2015)

\bibitem{DSS}
Song, S., Xiao, J.: Deep sliding shapes for amodal 3d object detection in rgb-d images. In: Proceedings of the IEEE conference on computer vision and pattern recognition. pp. 808--816 (2016)

\bibitem{Waymo}
Sun, P., Kretzschmar, H., Dotiwalla, X., Chouard, A., Patnaik, V., Tsui, P., Guo, J., Zhou, Y., Chai, Y., Caine, B., et~al.: Scalability in perception for autonomous driving: Waymo open dataset. In: Proceedings of the IEEE/CVF conference on computer vision and pattern recognition. pp. 2446--2454 (2020)

\bibitem{tang2019transferable}
Tang, Y.S., Lee, G.H.: Transferable semi-supervised 3d object detection from rgb-d data. In: Proceedings of the IEEE/CVF International Conference on Computer Vision. pp. 1931--1940 (2019)

\bibitem{BoxPC}
Tang, Y.S., Lee, G.H.: Transferable semi-supervised 3d object detection from rgb-d data. In: Proceedings of the IEEE/CVF International Conference on Computer Vision. pp. 1931--1940 (2019)

\bibitem{Weankmono3d}
Tao, R., Han, W., Qiu, Z., Xu, C.z., Shen, J.: Weakly supervised monocular 3d object detection using multi-view projection and direction consistency. In: Proceedings of the IEEE/CVF Conference on Computer Vision and Pattern Recognition. pp. 17482--17492 (2023)

\bibitem{gemini}
Team, G., Anil, R., Borgeaud, S., Wu, Y., Alayrac, J.B., Yu, J., Soricut, R., Schalkwyk, J., Dai, A.M., Hauth, A., et~al.: Gemini: a family of highly capable multimodal models. arXiv preprint arXiv:2312.11805  (2023)

\bibitem{FCOS}
Tian, Z., Shen, C., Chen, H., He, T.: Fcos: Fully convolutional one-stage object detection. In: Proceedings of the IEEE/CVF international conference on computer vision. pp. 9627--9636 (2019)

\bibitem{llama}
Touvron, H., Lavril, T., Izacard, G., Martinet, X., Lachaux, M.A., Lacroix, T., Rozi{\`e}re, B., Goyal, N., Hambro, E., Azhar, F., et~al.: Llama: Open and efficient foundation language models. arXiv preprint arXiv:2302.13971  (2023)

\bibitem{PGD}
Wang, T., Xinge, Z., Pang, J., Lin, D.: Probabilistic and geometric depth: Detecting objects in perspective. In: Conference on Robot Learning. pp. 1475--1485. PMLR (2022)

\bibitem{lsmol}
Wang, Y., Chen, Y., ZHANG, Z.X.: 4d unsupervised object discovery. Advances in Neural Information Processing Systems  \textbf{35},  35563--35575 (2022)

\bibitem{world_model_yuqi}
Wang, Y., He, J., Fan, L., Li, H., Chen, Y., Zhang, Z.: Driving into the future: Multiview visual forecasting and planning with world model for autonomous driving. In: Proceedings of the IEEE/CVF Conference on Computer Vision and Pattern Recognition. pp. 14749--14759 (2024)

\bibitem{FGR}
Wei, Y., Su, S., Lu, J., Zhou, J.: Fgr: Frustum-aware geometric reasoning for weakly supervised 3d vehicle detection. In: 2021 IEEE International Conference on Robotics and Automation (ICRA). pp. 4348--4354. IEEE (2021)

\bibitem{Mlcvnet}
Xie, Q., Lai, Y.K., Wu, J., Wang, Z., Zhang, Y., Xu, K., Wang, J.: Mlcvnet: Multi-level context votenet for 3d object detection. In: Proceedings of the IEEE/CVF conference on computer vision and pattern recognition. pp. 10447--10456 (2020)

\bibitem{Second}
Yan, Y., Mao, Y., Li, B.: Second: Sparsely embedded convolutional detection. Sensors  \textbf{18}(10), ~3337 (2018)

\bibitem{Centerpoint}
Yin, T., Zhou, X., Krahenbuhl, P.: Center-based 3d object detection and tracking. In: Proceedings of the IEEE/CVF conference on computer vision and pattern recognition. pp. 11784--11793 (2021)

\bibitem{AutolabelSDF}
Zakharov, S., Kehl, W., Bhargava, A., Gaidon, A.: Autolabeling 3d objects with differentiable rendering of sdf shape priors. In: Proceedings of the IEEE/CVF Conference on Computer Vision and Pattern Recognition. pp. 12224--12233 (2020)

\bibitem{zakharov2020autolabeling}
Zakharov, S., Kehl, W., Bhargava, A., Gaidon, A.: Autolabeling 3d objects with differentiable rendering of sdf shape priors. In: Proceedings of the IEEE/CVF Conference on Computer Vision and Pattern Recognition. pp. 12224--12233 (2020)

\bibitem{voxel_mamba}
Zhang, G., Fan, L., He, C., Lei, Z., Zhang, Z., Zhang, L.: Voxel mamba: Group-free state space models for point cloud based 3d object detection. arXiv preprint arXiv:2406.10700  (2024)

\bibitem{monodetr}
Zhang, R., Qiu, H., Wang, T., Guo, Z., Cui, Z., Qiao, Y., Li, H., Gao, P.: Monodetr: Depth-guided transformer for monocular 3d object detection. In: Proceedings of the IEEE/CVF International Conference on Computer Vision. pp. 9155--9166 (2023)

\bibitem{H3dnet}
Zhang, Z., Sun, B., Yang, H., Huang, Q.: H3dnet: 3d object detection using hybrid geometric primitives. In: Computer Vision--ECCV 2020: 16th European Conference, Glasgow, UK, August 23--28, 2020, Proceedings, Part XII 16. pp. 311--329. Springer (2020)

\bibitem{SUN-RGBD}
Zhou, B., Lapedriza, A., Xiao, J., Torralba, A., Oliva, A.: Learning deep features for scene recognition using places database. Advances in neural information processing systems  \textbf{27} (2014)

\bibitem{Voxelnet}
Zhou, Y., Tuzel, O.: Voxelnet: End-to-end learning for point cloud based 3d object detection. In: Proceedings of the IEEE conference on computer vision and pattern recognition. pp. 4490--4499 (2018)

\end{thebibliography}
\end{document}